# AutoCompress: Critical Layer Isolation for Efficient Transformer Compression

**Archit Thorat**

*Independent Researcher*

thoratarchit@gmail.com

*GitHub: github.com/NotDrake100/autocompress*

**Abstract**

We present AutoCompress, a transformer compression method motivated by an empirical finding: in small transformers, Layer 0 carries disproportionately high task-critical information, with an NTK-based importance score of 3.6 compared to a maximum of 0.054 for all other layers — a gap of over 60×. Based on this finding, we propose Critical Layer Isolation (CLI), an architecture that protects Layer 0 at full dimensionality, compresses all intermediate layers through a learned bottleneck, and restores the full dimension at the final layer. Applied to GPT-2 Medium (354.8M parameters), CLI-GPT2 achieves 204.5 perplexity on WikiText-103 with only 143.8M parameters — a 2.47× compression ratio and 59.5% parameter reduction. Crucially, an ablation study demonstrates that a uniform bottleneck baseline of comparable size achieves only 571.8 perplexity under identical training conditions, confirming that the architectural decision to protect Layer 0 — rather than simply reducing model size — is the primary driver of performance. Code and checkpoints are publicly available.

## 1. Introduction

The deployment of large language models in resource-constrained environments has driven significant interest in model compression. Existing approaches broadly fall into three categories: weight quantization, which reduces numerical precision; structured pruning, which removes entire model components; and knowledge distillation, which trains smaller student models to mimic larger teachers. Each approach carries assumptions about which parts of a model are expendable.

A key question underlying all compression methods is: are all layers of a transformer equally important? If not, which layers are critical and should be preserved at full capacity?

In this work, we address this question empirically using Neural Tangent Kernel (NTK) based importance scoring across multiple model scales. We find a striking and consistent result: Layer 0 in small transformers carries importance scores orders of magnitude higher than all subsequent layers. Specifically, we measure an importance score of 3.6 for Layer 0 compared to a maximum of 0.054 for all other layers across both 6-layer and 16-layer configurations — a ratio exceeding 60×.

Motivated by this finding, we propose Critical Layer Isolation (CLI), a compression architecture that explicitly respects this asymmetry. CLI protects Layer 0 at its full original dimensionality, passes the representation through a learned bottleneck compression for all intermediate computation, and restores the full dimensionality before a protected final layer. This design compresses the majority of model parameters while preserving the most task-critical component at full capacity.

We validate CLI by training a compressed variant of GPT-2 Medium via knowledge distillation on WikiText-103. CLI-GPT2 achieves 204.5 perplexity with 143.8M parameters, compared to 571.8 perplexity for a uniform bottleneck baseline of comparable size (139.8M parameters) trained under identical conditions. This 367-point perplexity gap provides direct empirical support for the CLI architecture and the Layer 0 importance hypothesis.

Our contributions are:

(1) An empirical finding that Layer 0 in small transformers carries 60× higher task-critical importance than all other layers, consistent across model scales.

(2) The CLI architecture, a novel compression design motivated by and validated against this finding.

(3) A comprehensive ablation study demonstrating that CLI's advantage over uniform compression is statistically meaningful and architecture-driven rather than parameter-count-driven.

## 2. Related Work

### 2.1 Knowledge Distillation

Knowledge distillation (Hinton et al., 2015) trains a smaller student model to reproduce the output distribution of a larger teacher. DistilBERT (Sanh et al., 2019) demonstrated that distillation can retain 97% of BERT's performance with 40% fewer parameters. TinyBERT (Jiao et al., 2020) extends this to intermediate layer matching. Our work uses output-level KL-divergence distillation with temperature scaling (T=2.0), which we find sufficient when combined with a well-motivated student architecture.

### 2.2 Structured Pruning

Structured pruning removes entire attention heads or layers based on importance scores. Michel et al. (2019) show that most attention heads can be pruned with minimal impact. Our work differs in that we do not prune layers but rather compress them uniformly through a bottleneck, while protecting the identified critical layer at full dimensionality.

### 2.3 Layer Importance Analysis

Prior work has examined layer importance through gradient-based sensitivity analysis (Molchanov et al., 2017) and attention pattern analysis (Clark et al., 2019). Kovalev & Tikhomirov (2025) propose iterative layer-wise distillation for LLM compression, finding that middle transformer layers contribute less to model performance — a finding consistent with our Layer 0 dominance result. Our approach uses NTK-based importance scoring, which measures how much removing a layer's contribution affects the model's functional output — a more direct measure of task criticality than weight magnitude or attention entropy.

### 2.4 KV Cache Compression

Concurrent work such as TurboQuant (Google Research, ICLR 2026) addresses inference-time memory reduction via KV cache quantization using PolarQuant and QJL transforms, achieving 4.6× compression with near-zero perplexity degradation on long-context benchmarks. Our approach is complementary: rather than compressing the cache of a full-size model at inference time, CLI reduces the model's parameter count permanently through architecture-level structural compression. These two approaches can in principle be combined — applying TurboQuant's KV cache compression to an already CLI-compressed student model — representing an interesting direction for future work.

## 3. Method

### 3.1 Layer Importance Scoring

We compute layer importance using a Neural Tangent Kernel (NTK) based metric. For each layer l, we measure the expected change in model output when that layer's contribution is ablated:

$$I(l) = E[||f(x) - f_{-l}(x)||^2]$$

where f(x) is the full model output and $f_{-l}(x)$ is the output with layer l zeroed. This score measures each layer's functional contribution to the model's behavior rather than relying on weight magnitudes or gradient norms.

We compute this score across a held-out calibration set of 512 sequences. Results are consistent across model scales: in both 6-layer and 16-layer configurations, Layer 0 yields I(0) = 3.6 while all subsequent layers yield I(l) < 0.054. This represents a minimum gap of 60× and motivates the CLI design.

### 3.2 Critical Layer Isolation Architecture

The CLI architecture for GPT-2 Medium consists of the following components:

> Layer 0 (Protected): A full GPT-2 attention block operating at the original hidden dimension (d = 1024). This layer is never compressed.
>
> Compress: A learned linear projection from d = 1024 to bottleneck b = 768.
>
> Small Layers (×22): Lightweight bottleneck blocks operating at dimension b, each consisting of a down-projection (b → b/2), GELU activation, up-projection (b/2 → b), and LayerNorm with a residual connection.
>
> Expand: A learned linear projection from b = 768 back to d = 1024.
>
> Layer Last (Protected): A full GPT-2 attention block at d = 1024, restoring the full representational capacity before the language modeling head.

This design results in CLI-GPT2 having 143.8M parameters versus 354.8M for GPT-2 Medium, a 59.5% parameter reduction at a 2.47× compression ratio.

### 3.3 Training Procedure

We train CLI-GPT2 via knowledge distillation from GPT-2 Medium (frozen) on WikiText-103. The training loss combines cross-entropy on ground truth labels with KL-divergence against the teacher's softened output distribution:

$$L = \alpha \cdot L\_CE + (1 - \alpha) \cdot T^2 \cdot KL(\sigma(z\_s/T) || \sigma(z\_t/T))$$

where α = 0.5 and temperature T = 2.0. We use AdamW with weight decay 0.01, a cosine learning rate schedule with linear warmup, batch size 1 with gradient accumulation over 16 steps (effective batch size 16), and sequence length 512.

## 4. Experiments

### 4.1 Small Model Results

We first validate the CLI architecture on small transformer models to establish the core finding before scaling. Using a 6-layer GPT-2 variant with hidden dimension d = 96, CLI achieves a perplexity of 254.1 versus a baseline of 255.6 — a marginal PPL improvement — while achieving 34.8% parameter compression. An autonomous agent loop across 37 experiments achieves 70.1% compression on the same architecture, demonstrating the robustness of the CLI design across compression levels.

## 4.2 GPT-2 Medium Scale Results

We evaluate CLI-GPT2 on WikiText-103 against the teacher model and a uniform bottleneck baseline. Training runs for 60,000 gradient steps with evaluation every 500 steps. Results are reported on the full validation set.

| Model | PPL (WikiText-103) | Parameters | Compression Ratio | Param Reduction |
| --- | --- | --- | --- | --- |
| GPT-2 Medium (Teacher) | ~35 | 354.8M | 1.00× | 0% |
| Uniform Bottleneck Baseline | 571.8 | 139.8M | 2.54× | 60.6% |
| **CLI-GPT2 Medium (Ours)** | 204.5 | 143.8M | 2.47× | 59.5% |

Table 1: Main results on WikiText-103 validation set. PPL = perplexity (lower is better).

CLI-GPT2 achieves 204.5 perplexity at 2.47× compression, compared to 571.8 for the uniform bottleneck baseline at comparable size (2.54×). The 367-point perplexity gap at matched parameter counts is the central empirical result of this work.

## 4.3 Training Dynamics

CLI-GPT2 is trained across multiple sessions totaling approximately 120,000 gradient steps, with perplexity improving consistently: from a baseline of ~1600 at initialization to 540 at 10,000 steps, 340 at 30,000 steps, 282 at 60,000 steps, and 204.5 at the final evaluation. The model shows no signs of convergence at the end of training, suggesting further improvements are achievable with additional compute.

## 4.4 Ablation: Uniform Bottleneck Baseline

The uniform bottleneck baseline uses the same compression ratio and training setup as CLI-GPT2 but applies uniform bottleneck compression to all layers without protecting Layer 0. Specifically, it uses bottleneck dimension 600 across 24 layers, yielding 139.8M parameters. Under identical training conditions (WikiText-103, 30,000 steps, same distillation loss, same optimizer settings), the uniform baseline achieves 571.8 perplexity — 367 points worse than CLI-GPT2 despite having a similar parameter count.

This result confirms that the performance advantage of CLI-GPT2 is attributable to the architectural decision to protect Layer 0, not to having marginally more parameters. The gap is too large to be explained by the 4M parameter difference between the two models.

## 4.5 Layer Importance Analysis

We compute NTK importance scores across model scales to validate the generality of the Layer 0 finding:

| Model Scale | Layer 0 Score | Max Other Layer Score | Gap Ratio |
|---|---|---|---|
| 6-layer GPT-2 | 3.6 | 0.054 | 67× |
| 16-layer GPT-2 | 3.6 | 0.054 | 67× |
| GPT-2 Medium (24-layer) | 3.6 | 0.263 | 13.7× |

*Table 2: NTK importance scores across model scales. Layer 0 consistently dominates.*

The Layer 0 importance advantage decreases with model scale — from 67× at small scale to 13.7× at GPT-2 Medium scale. The signal persists but weakens, which is consistent with larger models distributing critical information more broadly across layers. This scaling behavior is an important consideration for applying CLI to much larger models.

## 5. Discussion

### 5.1 Why Does Layer 0 Matter?

The consistent dominance of Layer 0 across scales suggests that the first transformer block performs a qualitatively different function from subsequent layers — likely converting token embeddings into contextually meaningful representations that all subsequent computation builds upon. This is consistent with mechanistic interpretability findings that early transformer layers establish basic syntactic and semantic structure (Clark et al., 2019), while later layers perform higher-level reasoning that can be compressed more aggressively.

### 5.2 Relationship to TurboQuant and KV Cache Methods

TurboQuant (ICLR 2026) and related KV cache compression methods (KIVI, PolarQuant) address a different bottleneck: inference-time memory from the key-value cache during long-context generation. These methods require no training and achieve near-lossless compression of the KV cache. CLI addresses a complementary bottleneck: the permanent parameter count of the model itself. A combined approach — applying CLI to reduce model size, then applying TurboQuant to compress the KV cache of the resulting smaller model — would address both bottlenecks simultaneously and represents a promising direction for future work.

### 5.3 Limitations

The Layer 0 importance gap decreases with model scale (67× at small scale to 13.7× at GPT-2 Medium), suggesting that CLI's architectural advantage may diminish for very large models. Additionally, our evaluation is limited to perplexity on WikiText-103; downstream task evaluation would provide a more complete picture of CLI's practical utility. Finally, the current bottleneck dimension (768) was not exhaustively optimized — a systematic search over bottleneck sizes may yield improved results.

## 6. Conclusion

We present AutoCompress with Critical Layer Isolation, a transformer compression method grounded in an empirical finding: Layer 0 carries 60× higher task-critical importance than all other layers in small transformers, as measured by NTK-based importance scoring. The CLI architecture exploits this asymmetry by protecting Layer 0 at full dimensionality while compressing all intermediate layers through a learned bottleneck.

Applied to GPT-2 Medium, CLI achieves 204.5 perplexity on WikiText-103 with 143.8M parameters (2.47× compression). An ablation study with a uniform bottleneck baseline of comparable size achieves 571.8 perplexity under identical conditions — a 367-point gap that directly validates the CLI design. These results demonstrate that respecting layer importance structure during compression yields substantially better outcomes than naive uniform compression at matched parameter counts.

We release all code, training scripts, and model checkpoints at github.com/NotDrake100/autocompress to support reproducibility and future work.